\documentclass{article}

\usepackage{arxiv}

\usepackage[utf8]{inputenc} % allow utf-8 input
\usepackage[T1]{fontenc}    % use 8-bit T1 fonts
\usepackage{hyperref}       % hyperlinks
\usepackage{url}            % simple URL typesetting
\usepackage{booktabs}       % professional-quality tables
\usepackage{amsfonts}       % blackboard math symbols
\usepackage{nicefrac}       % compact symbols for 1/2, etc.
\usepackage{microtype}      % microtypography
\usepackage{lipsum}
\usepackage{graphicx}
\usepackage{orcidlink}
\graphicspath{ {./images/} }

\usepackage{tabularx}
\usepackage{siunitx}
\usepackage{placeins}
\usepackage{multirow}
\usepackage{xcolor,colortbl}
\usepackage{amsmath}
\usepackage{amssymb}
\usepackage{makecell}
\usepackage{wrapfig}
\usepackage[normalem]{ulem}
\usepackage{svg}
\usepackage{pifont}
\usepackage{enumitem}
\usepackage[accsupp]{axessibility}  % Improves PDF readability for those with disabilities.

\newcommand{\etal}{et al.\@\xspace}
\usepackage{xspace}

\newcommand{\xmark}{\ding{55}}

\newcommand{\filledstar}{\ensuremath{\bigstar}}

\title{TinyDETR-Pose: Towards End-to-End Real-Time Single-Stage 6DoF Object Pose Estimation with Lightweight Transformers}

\author{
 Paul Julius Kühn\,\orcidlink{0000-0003-2458-0030} \\
  Fraunhofer IGD\\
  64283 Darmstadt, Germany \\
  \texttt{paul.julius.kuehn@igd.fraunhofer.de} \\
   \And
 Duc Anh Nguyen\,\orcidlink{0009-0003-5324-5482} \\
  Fraunhofer IGD, 64283 Darmstadt, Germany\\
  TU Darmstadt, 64289 Darmstadt, Germany \\
  \texttt{duc.anh.nguyen@stud.tu-darmstadt.de} \\
   \And
 Saptarshi Neil Sinha\,\orcidlink{0000-0001-6637-0379} \\
  Fraunhofer IGD\\
  64283 Darmstadt, Germany \\
  \texttt{saptarshi.neil.sinha@igd.fraunhofer.de} \\
  \AND
 Michael Weinmann\,\orcidlink{0000-0003-3634-0093} \\
  TU Delft\\
  2628 CD Delft, Netherlands \\
  \texttt{m.weinmann@tudelft.nl} \\
  \And
 Arjan Kuijper\,\orcidlink{0000-0002-6413-0061} \\
  Fraunhofer IGD, 64283 Darmstadt, Germany\\
  TU Darmstadt, 64289 Darmstadt, Germany \\
  \texttt{arjan.kuijper@igd.fraunhofer.de} \\
}

\begin{document}
\maketitle

\begin{abstract}
Real-time 6DoF object pose estimation on resource-constrained hardware remains
challenging, as accurate correspondence-based and refinement pipelines typically
rely on non-differentiable PnP/RANSAC stages or costly iterative refinement,
while recent foundation-model-based approaches incur inference costs that are
prohibitive for edge deployment. We present \mbox{\textbf{TinyDETR-Pose}}, a
lightweight, end-to-end, single-stage framework that jointly detects objects and
regresses their full 6D pose in a single forward pass. Built on the efficient
LW-DETR architecture, TinyDETR-Pose formulates detection and pose estimation as
a set-prediction problem and attaches dedicated MLP heads for rotation,
monocular depth, and projected object center regression to each decoder query,
eliminating the need for PnP, NMS (non-maximum suppression), or iterative pose refinement. 
Object symmetries are handled through a ADD-S loss
applied uniformly to all objects, without the need for object-specific
loss schedules or separate geodesic/ADD supervision. In addition, predictions
are assigned to ground truth using a symmetry-safe Hungarian matcher based on
class and 2D spatial cues, yielding stable assignment under symmetry and depth
ambiguity. On YCB-V, TinyDETR-Pose achieves a comparable ADD-S AUC of
\textbf{85.9}, while requiring up to $\mathbf{72.7 \%}$ fewer
parameters than other DETR-based single-stage pose-estimation approaches.
Due to its compact design, TinyDETR-Pose runs in real time and achieves an
inference latency of only ${\sim}\mathbf{4.5}$\,ms per frame on an NVIDIA
Jetson Nano using TensorRT, demonstrating that accurate end-to-end
transformer-based 6D pose estimation can be made practical for edge deployment.

\end{abstract}
\section{Introduction}\label{sec:introduction}
Accurately estimating an object's 6DoF (six degrees of freedom) pose, meaning
its full 3D rotation and translation relative to the camera, is essential in
various fields such as robotics and augmented
reality~\cite{tremblay2018deep, marchand2016pose}. Despite remarkable progress,
there is still a basic trade-off between the accuracy and efficiency of these
methods, which makes it challenging to deploy them in real-time
systems~\cite{6dSurvey}. The baseline methods rely on multi-stage pipelines that
establish dense 2D--3D correspondences and recover the pose via RANSAC-based PnP
solvers~\cite{pvnet, dpod, pix2pose, epnp}, making them non-differentiable and
expensive. Direct regression approaches~\cite{posecnn, gdrnet, sopose, cdpn}
bypass PnP but still depend on a separate 2D detector, preserving a two-stage
structure. Foundation models~\cite{foundationposewen2024, megapose, gigaPose,
ornek2024foundpose} achieve impressive zero-shot generalization, but ttheir inference times are unsuitable for real-time systems; for example, FoundationPose~\cite{foundationposewen2024} requires ${\sim}1.3$\,s per
object on high-end desktop GPUs, precluding edge deployment. This performance
limitation motivates single-stage architectures that jointly detect all objects
and estimate their 6D poses in a single forward
pass~\cite{yolox6d, poet, T6DDirect, YOLOPoseV2}. However, existing
single-stage methods struggle with monocular depth ambiguity, lack principled
symmetry handling, and provide limited rotation supervision.

\noindent We present \textbf{TinyDETR-Pose}, an end-to-end, single-stage 6D
pose estimation framework built on the lightweight detection transformer
LW-DETR~\cite{chen2024lwdetr} compatible with edge devices. We treat joint detection and pose estimation as a
set-prediction problem~\cite{carion2020detr,T6DDirect}, regressing each
instance's full 6D pose directly from the decoder output via dedicated MLP heads
for rotation, monocular depth, and projected object center---achieving 
inference in 4.5\,ms on Jetson Nano.\\
\noindent Our main contributions are:\\
- A single-stage, query-based architecture that jointly detects objects and regresses their full 6D pose without PnP, NMS, or iterative refinement.\\
- A unified ADD-S training objective that supervises the complete pose without requiring geodesic losses or per-object loss curricula.\\
-  A projected-center-based bipartite matcher that excludes rotation and
depth from the assignment cost, ensuring stable matching under symmetry
and depth ambiguity.\\
\noindent We evaluate on YCB-V~\cite{posecnn}, comparing against
single-stage methods~\cite{yolox6d, poet, T6DDirect, YOLOPoseV2} and
direct-regression baselines~\cite{gdrnet, posecnn}.

\begin{figure}[htb!]
  \centering
  \includegraphics[width=\textwidth]{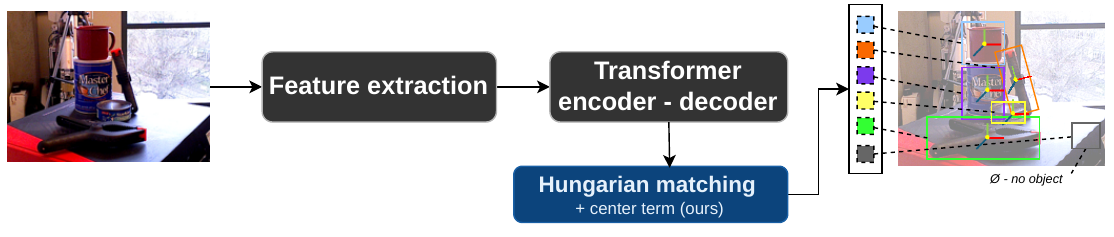}
  \caption{\textbf{Overview of TinyDETR-Pose.} Given a single RGB image, our model jointly detects all objects and regresses their full 6D pose in one forward pass via set prediction.}
  \label{fig:fig:lwdetr6d_overview}
\end{figure}
\section{Related Work}\label{sec:related}
Deep learning approaches to 6D object pose estimation broadly fall into three
categories: indirect correspondence-based methods, direct regression frameworks,
and foundation models. Orthogonally, DETR-style set-prediction architectures
have been extended beyond 2D detection to end-to-end 6D pose estimation,
motivating our query-based formulation.

\subsection{Indirect and Direct Regression Methods}
\textbf{Indirect methods} establish 2D--3D correspondences and recover the pose
via PnP~\cite{epnp}. BB8~\cite{bb8} regresses projections of bounding-box
corners; PVNet~\cite{pvnet} improves occlusion robustness through pixel-wise
voting with RANSAC~\cite{RANSAC}; DPOD \cite{dpod} and
Pix2Pose~\cite{pix2pose} extend this to dense UV and per-pixel 3D coordinate
maps, respectively. While robust, these pipelines remain multi-stage and are typically not end-to-end differentiable due to the PnP/RANSAC step. As a result, intermediate prediction errors can propagate to the final pose, the correspondence estimator cannot be trained directly from the final pose objective, and the overall pipeline depends on additional hand-tuned components such as RANSAC thresholds.\\
\textbf{Direct methods} regress object rotation and translation directly from the
image. PoseCNN~\cite{posecnn} regressed quaternions but suffered from
representation discontinuities later addressed by the continuous 6D
parameterization of Zhou et al.~\cite{rotation6d}. CDPN~\cite{cdpn}
disentangles rotation and translation into separate branches, and
GDR-Net~\cite{gdrnet} bridges direct and indirect paradigms by guiding
regression with dense correspondence maps. SO-Pose~\cite{sopose} adds
self-occlusion cues for complementary supervision, while
CosyPose~\cite{cosypose} extends the paradigm to multi-object scenes via
iterative render-and-compare refinement, achieving top results on the BOP
benchmark~\cite{hodan2018bopbenchmark6dobject}. Despite their accuracy, most
direct methods still rely on a separate 2D detector~\cite{gdrnet, cdpn},
making them two-stage at inference.

\subsection{Foundation Models for Unseen Objects}
Recent zero-shot 6D pose estimation methods aim to improve generalization by
leveraging synthetic rendering, template matching, and foundation-model features.
MegaPose~\cite{megapose} introduced a render-and-compare strategy for zero-shot
pose estimation but requires over 15\,s per object.
GigaPose~\cite{gigaPose} achieves a 35$\times$ speedup through discriminative
templates, yet still needs several hundred milliseconds.
FoundationPose~\cite{foundationposewen2024} combines large-scale synthetic
training with neural implicit representations for state-of-the-art BOP
results~\cite{hodan2018bopbenchmark6dobject}, and
FoundPose~\cite{ornek2024foundpose} shows that frozen DINOv2~\cite{oquab2023dinov2}
features suffice for synthetic-to-real matching without task-specific training.
Despite their generalization capabilities, these models are prohibitively heavy
for edge devices; for example FoundationPose requires ${\sim}1.3$\,s per object even on
high-end desktop GPUs.

\subsection{Single-Stage and Transformer-based Approaches}
Single-stage architectures unify detection and pose regression in one forward
pass. YOLO-6D-Pose~\cite{yolox6d} extends YOLOX~\cite{yolox2021} to infer all
poses simultaneously. EfficientPose~\cite{bukschat2020efficientposeefficientaccuratescalable}
adapts EfficientDet~\cite{efficientdetscalableefficientobject} for scalable
end-to-end estimation. Transformer-based variants further improve accuracy
through global context: T6D-Direct~\cite{T6DDirect} treats multi-object pose
estimation as set prediction, PoET~\cite{poet} leverages multi-scale features
from pre-trained detectors, and YOLOPose~\cite{YOLOPoseV1, YOLOPoseV2} combines keypoint
regression with a learnable rotation module to eliminate non-differentiable PnP
solvers at real-time speed. However, such transformer-based designs typically
require substantially larger models, with approximately $3.5\times$ more
parameters than our approach. More recently, RACE-6D~\cite{ha2026race6d} pushes
this paradigm further by predicting detections and poses for cluttered
multi-object scenes in a single transformer forward pass, combining
keypoint-conditioned rotation refinement, bounding-box normalization for scale
invariance, and proportional depth refinement to reach competitive accuracy at up
to $5\times$ lower latency than prior pipelines. Our work differs in three respects: 
\emph{(i)} it uses no object-corner or multi-keypoint supervision beyond the projected object center; \emph{(ii)} it uses a unified ADD-S pose loss without explicit symmetry labels and omits keypoint-conditioned refinement; and \emph{(iii)} it explicitly targets edge deployment.

\section{Methodology}\label{sec:methodology}
\begin{figure}[htb!]
  \centering
  \includegraphics[width=\textwidth]{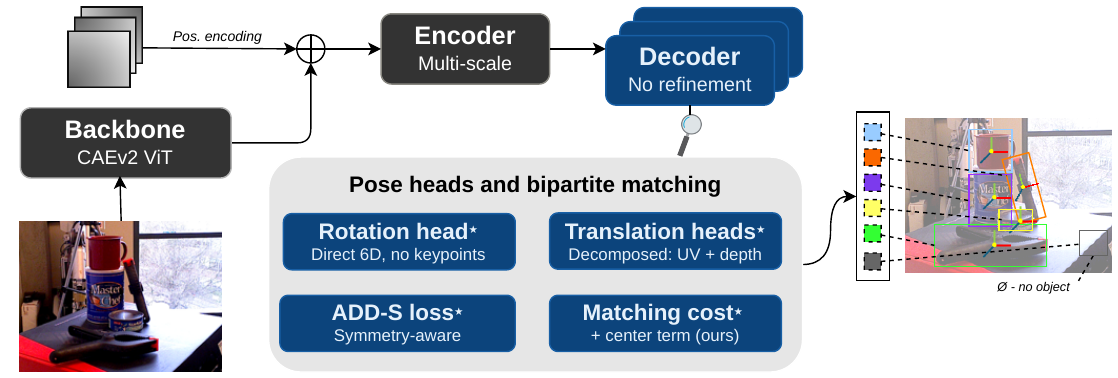}
    \caption{\textbf{Architecture of TinyDETR-Pose.} A CAEv2-pretrained ViT
  backbone~\cite{caev2} feeds multi-scale features into a 3-layer transformer decoder operating on $N$ learnable object queries. Hungarian
  matching~\cite{Hungarian} assigns predictions to ground truth during training,
  using our projected object-center cost (Sec.~\ref{subsec:matcher}). Each matched
  query is decoded by shared, parallel MLP heads for class, box, rotation, and
  translation, see Sec.~\ref{subsec:implementation} for details on the translation decomposition and rotation parameterization.\\  
  $\filledstar$ \textit{differs from T6D-Direct~\cite{T6DDirect} / RACE-6D~\cite{ha2026race6d}.}}
  \label{fig:lwdetr6d_detail}
\end{figure}

In this section, we present \textbf{TinyDETR-Pose}, an end-to-end 6D object pose
estimation pipeline suitable for edge devices. While our approach builds on LW-DETR~\cite{chen2024lwdetr}, we introduce three
methodological extensions for 6D object pose estimation:
(i)~dedicated pose prediction heads for translation and rotation estimation,
(ii)~a pose-aware loss formulation for supervising 6D pose predictions, and
(iii)~a 2D center-based matching strategy that incorporates the projected object
center $(u,v)$ into the Hungarian assignment to improve pose-query association
while remaining robust to symmetry and depth ambiguity. An architectural overview
is provided in Fig.~\ref{fig:lwdetr6d_detail}.
\subsection{6D Object Pose Estimation as Set Prediction}\label{subsec:matcher}
Inspired by the set-based prediction paradigm of DETR~\cite{carion2020detr} and its extension to 6DoF pose estimation in T6D-Direct~\cite{T6DDirect}, we jointly regress 2D object detections and their corresponding 6D poses in a single forward pass. Building upon LW-DETR~\cite{chen2024lwdetr}, our method inherits its lightweight design, making it particularly suited for edge deployment. In contrast to prior transformer-based pose estimation methods~\cite{T6DDirect, poet} that regress the full translation vector directly, we decompose translation estimation into two distinct sub-problems~\cite{splitTrans}:
instead of directly regressing the full metric translation vector, we estimate
the projected object center in normalized \emph{UV} space and the depth $t_z$
separately. The lateral translation components $(t_x,t_y)$ are then recovered
using the camera intrinsic matrix $\mathbf{K}$. This factorization separates
image-plane localization from monocular depth estimation, reducing the coupling
between lateral position and scale/depth and leading to a more stable and
camera-aware regression problem.\\
\noindent\textbf{Set Prediction.} Given an input image of size ($H$$\times$$W$$\times$$3$), our model predicts a fixed-size set of 
$N$ tuples $\{\hat{q}_i\}_{i=1}^{N}$. Each tuple is represented as $\hat{q}_i = \bigl(\gamma_i,\, \mathbf{b}_i,\, \mathbf{uv}_{\mathrm{cntr}_i},\, t_{z_i},\, \mathbf{r}_{6\mathrm{D}_i}\bigr)$ comprising class label probabilities ($\gamma_i$), a bounding box ($\mathbf{b}_i$), and a 6DoF object pose split into translation ($\mathbf{uv}_{\mathrm{cntr}_i}$, $t_{z_i}$) and a 6D rotation vector ($\mathbf{r}_{6\mathrm{D}_i}$). Bounding boxes and projected centers are normalized to [0,1]. Rotation is represented by an unconstrained continuous 6D vector, and depth is predicted in positive metric units. For rotation, we adopt the continuous 6D representation of Zhou~\etal\cite{rotation6d}, which has been shown to yield superior training stability and downstream performance. Following the set prediction formulation of DETR~\cite{carion2020detr}, the decoder operates on 
$N$ learnable object queries, where $N$ is a hyperparameter set to exceed the maximum number of objects expected in any given image. Predictions not matched to a ground-truth target are supervised against a dedicated $\varnothing$ (\emph{no-object}) class, enabling the model to explicitly learn to suppress false positives.\\   
\textbf{Enhanced Bipartite Matching.}
Following DETR~\cite{carion2020detr}, we use Hungarian matching to assign each
predicted query $\hat{q}_i$ to a ground-truth instance $y_j$ by minimizing a
pairwise cost matrix $\mathcal{C}\!\left(\hat{q}_i, y_j\right)$ composed of
class confidence ($\mathcal{C}_{\text{cls}}$), bounding-box $\ell_1$ distance
($\mathcal{C}_{\text{bbox}}$), and IoU ($\mathcal{C}_{\text{IoU}}$). We
additionally introduce a center cost ($\mathcal{C}_{\text{cntr}}$), defined as the
$\ell_1$ distance between predicted and ground-truth 3D object centers projected
onto the image plane.

\noindent\textbf{Final matching cost.}
We define our matching cost as:
\begin{equation}
    \mathcal{C}\!\left(\hat{q}_i, y_j\right)
    = \lambda^{\mathcal{C}}_{\mathrm{cls}}\,\mathcal{C}_{\mathrm{cls}}
    + \lambda^{\mathcal{C}}_{\mathrm{bbox}}\,\mathcal{C}_{\mathrm{bbox}}
    + \lambda^{\mathcal{C}}_{\mathrm{IoU}}\,\mathcal{C}_{\mathrm{IoU}}
    + \lambda^{\mathcal{C}}_{\mathrm{cntr}}\,\mathcal{C}_{\mathrm{cntr}},
    \label{eq:final_cost}
\end{equation}
Because rotation and depth are inherently ambiguous, we supervise them only through the post-matching loss, with rotation and depth are excluded from matching and supervised after assignment; ADD-S provides symmetry-tolerant rotation supervision.
This design is consistent with the matching philosophy of DETR~\cite{carion2020detr} and its
successors~\cite{chen2024lwdetr, meng2021conditionaldetr, li2022dndetr,
liu2022dabdetr, zhao2024rtdetr, sun2021tsp}, which rely exclusively on 2D
geometric quantities in the assignment cost to ensure stable convergence,
deferring task-specific supervision entirely to the training loss.
We set the matching cost weights in $\mathcal{C}\!\left(\hat{q}_i, y_j\right)$ to:
\begin{equation}
    (\lambda^{\mathcal{C}}_{\text{cls}},\,
  \lambda^{\mathcal{C}}_{\text{bbox}},\,
  \lambda^{\mathcal{C}}_{\text{IoU}},\,
  \lambda^{\mathcal{C}}_{\text{cntr}}) 
= (2.0,\, 5.0,\, 2.0,\, 5.0)
\end{equation}
The $\lambda^{\mathcal{C}}_{\text{cls}}$, $\lambda^{\mathcal{C}}_{\text{bbox}}$, and $\lambda^{\mathcal{C}}_{\text{IoU}}$ weights follow the default matching
configuration of LW-DETR~\cite{chen2024lwdetr}. The center term is specific to our
pose-estimation formulation and was chosen empirically; setting
$\lambda^{\mathcal{C}}_{\text{cntr}}=5.0$ balances the center localization cost
with the bounding-box regression cost and yielded stable matching in our
experiments.
\subsection{Model Architecture \& Training Objective}\label{subsec:implementation}
\begin{figure}[htb!]
  \centering
  \includegraphics[width=\textwidth]{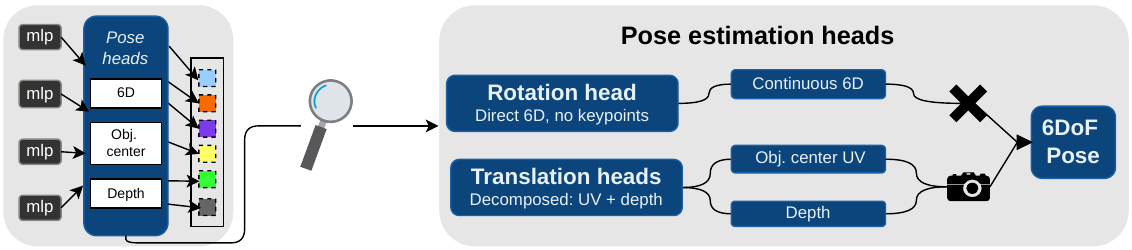}
\caption{\textbf{Pose estimation head.} Rotation is predicted in 6D~\cite{rotation6d} and orthogonalized to a $3\times3$ matrix. Translation follows~\cite{splitTrans}: the object center is predicted in normalized \textit{UV} space for Hungarian matching, while $t_z$ is estimated as depth. Back-projecting \textit{UV} and $t_z$ through $\mathbf{K}$ recovers the translation; combined with the predicted rotation, this yields the full 6DoF pose.}
\label{fig:lwdetr6d_poseheads}
\end{figure}

In the following, we detail the network architecture, the design of our pose
estimation head, and the training strategy including our proposed loss for 6DoF
object pose regression.\\
\textbf{Network architecture.}
Our pose estimation model builds upon the 2D detection framework
LW-DETR~\cite{chen2024lwdetr}. We retain its default configuration, including
the CAEv2~\cite{caev2} autoencoder backbone, the plain ViT encoder stack, the
projector structure, and a shallow 3-layer decoder, preserving its efficiency.
Specifically, we adopt the same number of encoder and decoder layers, attention
heads, hidden dimensions, and feed-forward network settings as LW-DETR. Our pose
estimation head comprises three MLPs that share the same input
dimension~($d_h$) and a single hidden layer, differing only in their output
dimensions, as detailed below.\\
\noindent\textbf{Pose estimation head design.}
All 6DoF pose parameters are predicted by three lightweight MLPs
(Fig.~\ref{fig:lwdetr6d_poseheads}), each with input dimension $d_h{=}256$ and a single hidden layer. The rotation head outputs a 6-dimensional vector using the continuous 6D parameterization~\cite{rotation6d}, from which the rotation matrix is recovered by orthogonalization.
For translation, following Simonelli et al.~\cite{splitTrans}, we decompose the
3D translation vector into two complementary subproblems instead of directly
regressing it in camera coordinates. The projected object center $(u,v)$ is
estimated in image space and supervised as a keypoint prediction task, while the
metric depth $t_z$ is treated as monocular depth estimation. This decomposition
separates image-plane localization from depth regression and allows each
subproblem to be supervised with an appropriate loss.
For each matched instance \(i\), let
\(\hat{\mathbf{c}}_i=(\hat{u}_i,\hat{v}_i)^\top\) denote the predicted projected
object center, and let \(\mathbf{c}_i=(u_i,v_i)^\top\) be the ground-truth center
obtained by projecting the 3D object center into the image using the camera
intrinsics~$\mathbf{K}$. 
The decoder produces $N$ predictions, but pose supervision is applied only to queries assigned to ground-truth objects by Hungarian matching. We therefore denote the number of matched prediction–target pairs by $N_{match}$ and average all pose-specific losses over these matched instances. We supervise the projected center using an OKS-based loss~\cite{yolox6d, YoloPose}:
\begin{equation}
\operatorname{OKS}_i =
\exp\!\left(
-\frac{\left\|\hat{\mathbf{c}}_i-\mathbf{c}_i\right\|_2^2}
{2s_i^2\sigma_{\mathrm{cntr}}^2}
\right),
\qquad
\mathcal{L}_{\mathrm{cntr}}
=
\frac{1}{N_{match}}
\sum_{i=1}^{N_{match}}
\left(1-\operatorname{OKS}_i\right),
\end{equation}
where \(s_i=\sqrt{w_i h_i}\) denotes the object scale from the ground-truth
bounding box and \(\sigma_{\mathrm{cntr}}\) controls the center-keypoint
tolerance.
For metric depth, we regress \(t_z\) in log-space:
\begin{equation}
e^z_i =
\log \hat{t}_{z,i} - \log t_{z,i}
=
\log\!\left(\frac{\hat{t}_{z,i}}{t_{z,i}}\right),
\end{equation}
\begin{equation}
\mathcal{L}_{t_z}
=
\frac{1}{N_{match}}
\sum_{i=1}^{N_{match}}
\operatorname{Smooth}_{\ell_1}^{\beta}\!\left(e^z_i\right),
\end{equation}
with
\begin{equation}
\operatorname{Smooth}_{\ell_1}^{\beta}(e)
=
\begin{cases}
\dfrac{1}{2\beta}e^2, & \text{if } |e| < \beta,\\[4pt]
|e|-\dfrac{\beta}{2}, & \text{otherwise}.
\end{cases}
\end{equation}
We set $\beta = 0.01$. At inference, the full translation $(t_x, t_y, t_z)$ is recovered by
back-projecting the predicted center $(u,v)$ through $\mathbf{K}^{-1}$ at the
predicted depth~$t_z$.\\
\noindent\textbf{Training strategy \& loss.} TinyDETR-Pose is a single-stage model that
jointly optimizes detection and 6D pose. For detection, we use an IoU-aware
classification loss $\mathcal{L}_{\text{cls}}$ (IA-BCE~\cite{cai2023aligndetr}),
an $\ell_1$ bounding-box loss $\mathcal{L}_{\text{bbox}}$, and a generalized IoU
loss $\mathcal{L}_{\text{IoU}}$~\cite{giou}. For pose estimation, we use the
center loss $\mathcal{L}_{\mathrm{cntr}}$, the log-depth loss
$\mathcal{L}_{t_z}$, and a geometry-grounded ADD-S loss. Rotation receives no
direct component-wise supervision; instead, it is optimized implicitly through
an ADD-S objective.
Because ADD-S measures the distance to the nearest transformed model point, it
provides a single pose-level signal for both symmetric and asymmetric objects
without requiring symmetry labels. Let
$\mathcal{M}_i=\{\mathbf{x}_{ij}\}_{j=1}^{M_i}$ denote the set of 3D model points
for object instance $i$; for each object mesh, we uniformly sample $512$ points
to construct $\mathcal{M}_i$. Given the predicted pose
$(\hat{\mathbf{R}}_i,\hat{\mathbf{t}}_i)$ and ground-truth pose
$(\mathbf{R}_i,\mathbf{t}_i)$, we define:
\[
\hat{\mathbf{p}}_{ij} = \hat{\mathbf{R}}_i \mathbf{x}_{ij} + \hat{\mathbf{t}}_i,
\qquad
\mathbf{p}_{ij} = \mathbf{R}_i \mathbf{x}_{ij} + \mathbf{t}_i .
\]

We use the symmetry-agnostic ADD-S distance for every instance:
\[
d_i =
\frac{1}{M_i}\sum_{k=1}^{M_i}
\min\limits_{j \in \{1,\dots,M_i\}}
\left\|
\hat{\mathbf{p}}_{ij} - \mathbf{p}_{ik}
\right\|_2 .
\]

The final ADD-S loss over all matched instances is:
\begin{equation}
\mathcal{L}_{\mathrm{ADD}\text{-}\mathrm{S}}
=
\frac{1}{N_{match}}
\sum_{i=1}^{N_{match}} d_i .
\label{eq:adds_loss}
\end{equation}

We use the following loss weight configuration: $\lambda^{\mathcal{L}}_{\text{cls}} = 1.0$, $\lambda^{\mathcal{L}}_{\text{bbox}} = 5.0$, $\lambda^{\mathcal{L}}_{\text{IoU}} = 2.0$, $\lambda^{\mathcal{L}}_{\text{cntr}} = 2.5$, $\lambda^{\mathcal{L}}_{\text{ADD-S}} = 30.0$, and $\lambda^{\mathcal{L}}_{t_z} = 15.0$. The weights were chosen such that the individual loss terms operate in roughly the same value range, while $\lambda^{\mathcal{L}}_{\text{ADD-S}}$ was determined through our ablation experiments (Sec.~\ref{subsec:ablation}).
Combining the loss criteria leads us to the following equation:
\begin{equation}
\label{eq:loss}
\begin{split}
\mathcal{L}_{\mathrm{total}} =\;
    & \lambda^{\mathcal{L}}_{\mathrm{cls}} \cdot \mathcal{L}_{\mathrm{cls}}
    + \lambda^{\mathcal{L}}_{\mathrm{bbox}} \cdot \mathcal{L}_{\mathrm{bbox}}
    + \lambda^{\mathcal{L}}_{\mathrm{IoU}} \cdot \mathcal{L}_{\mathrm{IoU}} \\
    & + \lambda^{\mathcal{L}}_{\mathrm{cntr}} \cdot \mathcal{L}_{\mathrm{cntr}}
    + \lambda^{\mathcal{L}}_{t_z} \cdot \mathcal{L}_{t_z}
    + \lambda^{\mathcal{L}}_{\mathrm{ADD}\text{-}\mathrm{S}} \cdot \mathcal{L}_{\mathrm{ADD}\text{-}\mathrm{S}} \\
\end{split}
\end{equation}
We use a ViT-Tiny initialized with CAEv2~\cite{caev2}, followed by LW-DETR~\cite{chen2024lwdetr} detector pretraining on Objects365~\cite{obj365} with a batch size of 32 for 60 epochs. We use AdamW~\cite{adamw} with a learning rate of $1e^{-4}$. $N$ is set to $100$.

\section{Experiments}
We describe the datasets and metrics, compare against state-of-the-art methods, and present ablation studies. Best results are \colorbox{yellow!50}{\textbf{highlighted}}.
\subsection{Setup}
We evaluate on YCB-V~\cite{posecnn}, which contains 21 household objects across
92 RGB-D sequences ($640\times480$). Following the standard protocol, we train
on the training sequences and 80K synthetic images and evaluate on 2,949
keyframes from the 12 test sequences~\cite{posecnn}. For the BOP evaluation
(Tab.~\ref{tab:ycbv_bop_metrics}), we use the official 900-image YCB-V
subset~\cite{hodan2018bop}. We report AUC for ADD(-S) and ADD-S, as well as the
official BOP average recall (AR) and its VSD, MSSD, and MSPD components.

\subsection{Comparison with state-of-the-art baselines}
We compare TinyDETR-Pose against representative methods across three tiers:
single-stage real-time methods, direct regression baselines, and
refinement-based upper bounds. Tab.~\ref{tab:ycbv_main} reports results
on the YCB-V benchmark~\cite{posecnn}.
Our method attains an AUC of ADD-S comparable to the state of the art while
using only $14.5\mathrm{M}$ parameters. In contrast, DETR-based approaches that
reach state-of-the-art accuracy rely on substantially larger parameter budgets,
primarily owing to their deeper decoder stacks and larger backbones. Despite
this reduction in model size, the gap between our model and the state of the art
on AUC of ADD-S remains small.
Notably, the CNN-based approach YOLO-6D-Pose (s)~\cite{yolox6d} achieves
state-of-the-art AUC of ADD(-S) with as few as $11.6\mathrm{M}$ parameters, yet
it relies on NMS as a hand-crafted post-processing step and is therefore not
truly end-to-end. Competing single-stage transformer-based methods, in turn,
typically require roughly $5\times$ more parameters to reach comparable
accuracy~\cite{YOLOPoseV1,YOLOPoseV2}. Our method departs from this trend: with
only $14.5\mathrm{M}$ parameters, it nearly matches the compactness of
CNN-based models while retaining the advantages of a fully end-to-end,
NMS-free transformer detector that directly predicts a set of poses. Its
accuracy on the AUC of ADD(-S) nonetheless leaves some headroom, which we
attribute to fine-grained pose precision and analyze in detail in our ablation
studies.
\begin{table*}[htb]
\centering
\resizebox{\textwidth}{!}{%
\begin{tabular}{lccccccc}
\toprule
\multirow{2}{*}{\textbf{Method}} &
\multirow{2}{*}{\textbf{RGB-only}} &
\multirow{2}{*}{\textbf{Refinement}} &
\multirow{2}{*}{\textbf{Real-time}} &
\multirow{2}{*}{\textbf{Params (M)}} &
\textbf{Num.} &
\textbf{AUC of} &
\textbf{AUC of} \\
& & & & &
\textbf{Decoders} &
\textbf{ADD(-S)} $\uparrow$ &
\textbf{ADD-S} $\uparrow$ \\
\midrule
\multicolumn{8}{l}{\textit{Refinement-based}} \\
\midrule
CosyPose (ECCV'20)~\cite{cosypose}
    & \checkmark & \checkmark & \xmark & -- & --
    & \cellcolor{yellow!50}\textbf{84.5}& 89.8 \\
DeepIM (ECCV'18)~\cite{deepim}
    & \checkmark & \checkmark & \xmark & -- & --
    & 81.9 & 88.1 \\
\midrule
\multicolumn{8}{l}{\textit{Direct regression}} \\
\midrule
GDR-Net (CVPR'21)~\cite{gdrnet}
    & \checkmark & \xmark & \checkmark & -- & --
    & 84.4 & 89.1 \\
SO-Pose (ICCV'21)~\cite{sopose}
    & \checkmark & \xmark & \checkmark & -- & --
    & 83.9 & 90.9 \\
PoseCNN (RSS'18)~\cite{posecnn}
    & \checkmark & \xmark & \checkmark & -- & --
    & 61.3 & 75.9 \\
\midrule
\multicolumn{8}{l}{\textit{Single-stage}} \\
\midrule
T6D-Direct$^\diamond$ (GCPR'21)~\cite{T6DDirect}
    & \checkmark & \xmark & \checkmark & -- & 6
    & 74.6 & 86.2 \\
PoET$^\diamond$ (CoRL'22)~\cite{poet}
    & \checkmark & \xmark & \checkmark & -- & 5
    & -- & 87.1 \\
YOLO-6D-Pose (s)(3DV'24)~\cite{yolox6d}
    & \checkmark & \xmark & \checkmark & 11.6 & --
    & 84.4 & -- \\
YOLOPoseV2$^\diamond$ (RAS'23)~\cite{YOLOPoseV2}
    & \checkmark & \xmark & \checkmark & 48.6 & 6
    & 82.6 & 90.1 \\
YOLOPoseV2-A$^\diamond$ (RAS'23)~\cite{YOLOPoseV2}
    & \checkmark & \xmark & \checkmark & 53.2 & 6
    & 83.3 & \cellcolor{yellow!50}\textbf{91.2} \\
\midrule
\rowcolor{gray!12}
\textbf{TinyDETR-Pose$^\diamond$ (Ours)}
    & \checkmark & \xmark & \checkmark & \textbf{14.5} & \textbf{3}
    & \textbf{64.6} & \textbf{85.9} \\
\bottomrule
\end{tabular}%
}
\caption{Comparison on YCB-V. $\diamond$ marks single-stage DETR-based methods;
decoder layers and parameter counts are reported where available. TinyDETR-Pose
achieves competitive ADD-S AUC with 14.5M parameters, 70.2--72.7\% fewer than
the DETR-based baselines with reported model sizes.}
\label{tab:ycbv_main}
\end{table*}
\noindent
\begin{table*}[t]
\centering
% Left side: discussion
\begin{minipage}[t]{0.44\textwidth}
\vspace{0pt}
\raggedright
\textbf{BOP-Metrics.}
RACE-6D achieves the highest reported AR of 76.7\% among the listed methods.
TinyDETR-Pose instead prioritizes efficiency,
reaching the lowest reported latency of 7\,ms---a reduction of
approximately 42\% over RACE-6D, 59\% over YOLO-6D-Pose, and 98\% over
CosyPose. This efficiency comes at the cost of lower recall, reflecting
a clear accuracy--latency trade-off.
\end{minipage}
\hfill
% Right side: table
\begin{minipage}[t]{0.55\textwidth}
\vspace{0pt}
\centering
\resizebox{\linewidth}{!}{%
\begin{tabular}{lccccc}
\toprule
\textbf{Method} &
\textbf{AR} &
$\mathbf{AR}_{\mathrm{VSD}}$ &
$\mathbf{AR}_{\mathrm{MSSD}}$ &
$\mathbf{AR}_{\mathrm{MSPD}}$ &
\textbf{ms} \\
\midrule
CosyPose (ECCV'20)~\cite{cosypose}
    & 65.5
    & --
    & --
    & --
    & 395 \\
YOLO-6D-Pose (s) (3DV'24)~\cite{yolox6d}
    & 70.8
    & 65.9
    & 72.3
    & 74.2
    & 16.9 \\
RACE-6D (CVPR'26)~\cite{ha2026race6d}
    & 76.7
    & 70.3
    & 79.9
    & 79.8
    & 12 \\
\midrule
\rowcolor{gray!12}
\textbf{TinyDETR-Pose$^*$ (Ours)}
    & \textbf{34.4}
    & \textbf{41.2}
    & \textbf{35.3}
    & \textbf{26.6}
    & \textbf{7} \\
\bottomrule
\end{tabular}%
}
\caption{Comparison on the YCB-V BOP benchmark using the official BOP
metrics (Average Recall \%). Higher is better for all AR metrics, while lower is better
for inference time. Runtime values are reported by the respective
methods. \textit{$^*$Results:} \url{https://bop.felk.cvut.cz/sub_info/41155/}}
\label{tab:ycbv_bop_metrics}
\end{minipage}
\end{table*} 

\begin{figure}[htb]
    \centering
    \setlength{\tabcolsep}{2pt}
    \begin{tabular}{ccc}
        \multicolumn{2}{c}{\textbf{Successful predictions}} &
        \textbf{Failure cases} \\

        \includegraphics[width=0.3\linewidth]{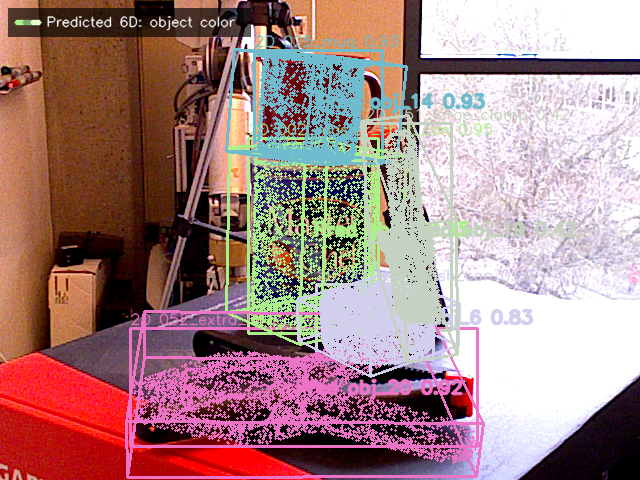} &
        \includegraphics[width=0.3\linewidth]{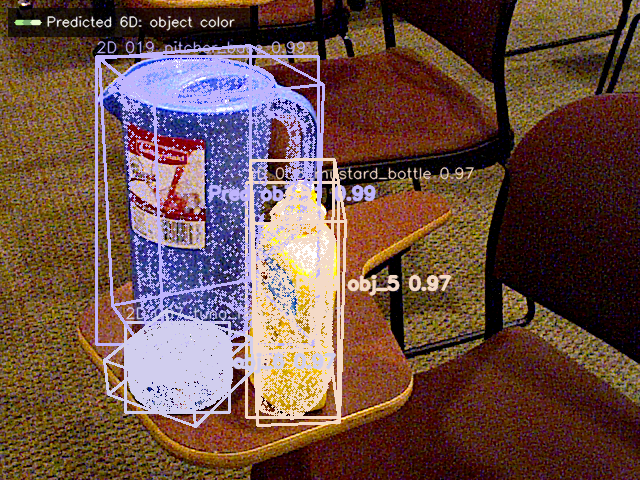} &
        \includegraphics[width=0.3\linewidth]{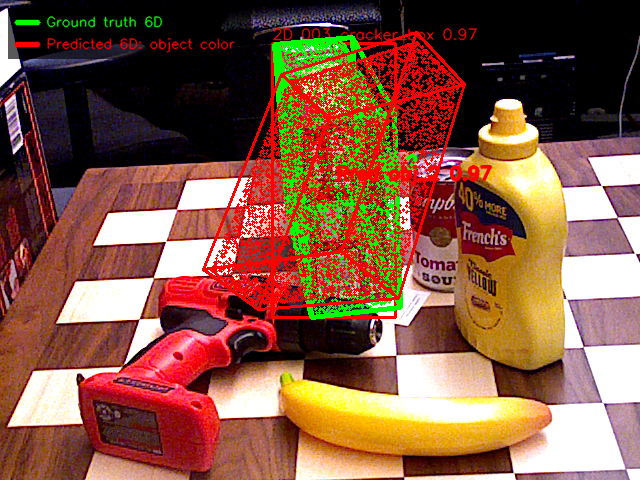} \\
        \small (a)  &
        \small (b) &
        \small (c)  \\[4pt]

        \includegraphics[width=0.3\linewidth]{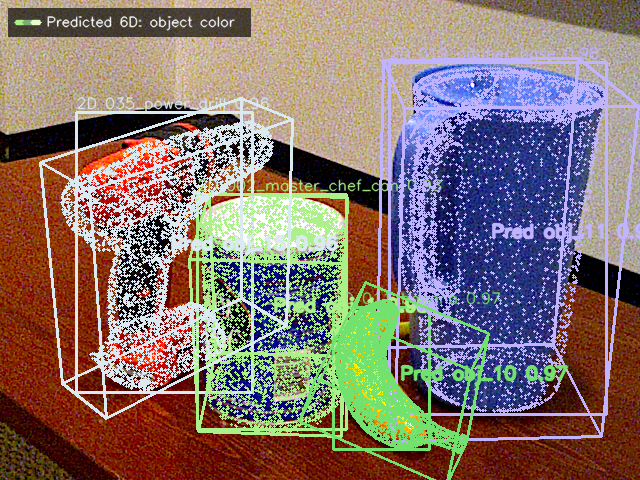} &
        \includegraphics[width=0.3\linewidth]{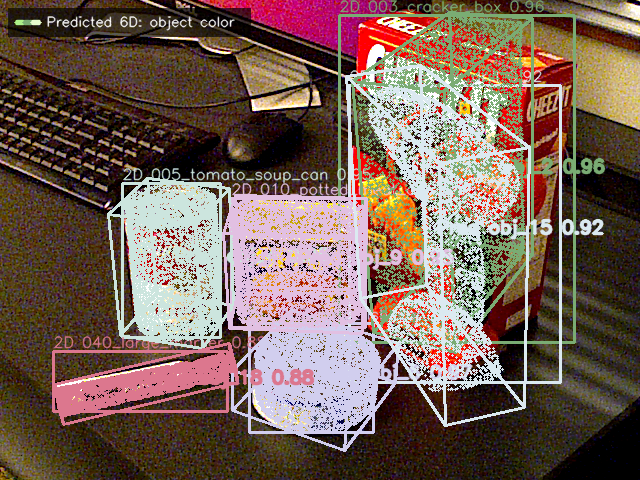} &
        \includegraphics[width=0.3\linewidth]{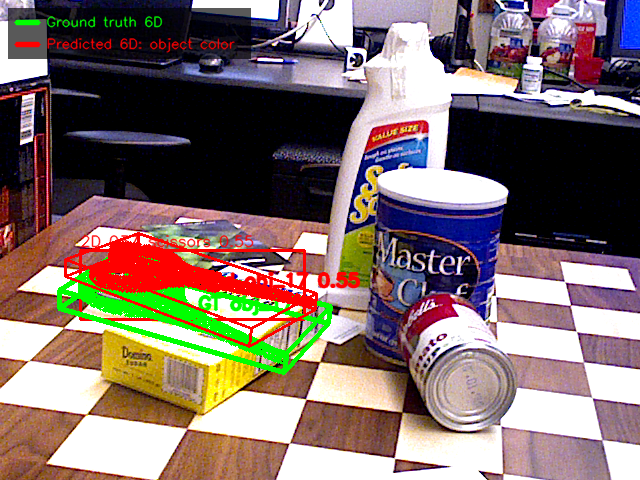} \\
        \small (d)  &
        \small (e)  &
        \small (f) 
    \end{tabular}
    \caption{
    Qualitative results of \textbf{TinyDETR-Pose} on YCB-V~\cite{posecnn}.
    Each row shows two successful predictions followed by one representative
    failure case. Failure cases: \colorbox{green!60}{Green} denotes ground-truth poses,
    and \colorbox{red}{red} denotes predictions from our model.
    The failure cases are mainly caused by strong occlusion.
    }
    \label{fig:qualitative_results}
\end{figure}

\noindent\textbf{Qualitative results.}
Fig.~\ref{fig:qualitative_results}~(a,b,d,e) shows successful predictions of
TinyDETR-Pose. The model aligns the predicted poses well with the
ground truth even in cluttered scenes and handles both small and large objects
reliably. Fig.~\ref{fig:qualitative_results}~(c,f) depicts representative
failure cases. We assume these errors are mainly caused by strong occlusion and
partial visibility, especially for the scissors, where large parts of the
object geometry are hidden and the pose becomes ambiguous.

\subsection{Ablation studies}~\label{subsec:ablation}
\noindent\textbf{Matcher analysis.} In the ablation of the projected object center matching cost $\mathcal{C}_{\mathrm{cntr}}$ (see Tab.~\ref{tab:ablation} (a)), we observe that adding the center term to the matching cost improves pose accuracy over matching on class and bounding-box cues alone.\\
\noindent\textbf{Effect of the ADD-S Loss Weight.}\label{subsec:abla}
Tab.~\ref{tab:ablation} (b) shows that the $\lambda^{\mathcal{L}}_{\mathrm{ADD}\text{-}\mathrm{S}}$ weight has a clear impact on pose-estimation performance. Increasing $\lambda^{\mathcal{L}}_{\mathrm{ADD}\text{-}\mathrm{S}}$ from $10.0$ to $20.0$ leads to a gradual improvement in both AUC of ADD-S and AUC of ADD(-S), indicating that stronger geometric supervision is beneficial. The best overall performance is obtained with $\lambda^{\mathcal{L}}_{\mathrm{ADD}\text{-}\mathrm{S}}=30.0$, achieving the highest AUC for both ADD-S and ADD(-S). Further increasing the coefficient to $50.0$ or $75.0$ degrades the AUC metrics, suggesting that an overly large ADD-S loss weight disturbs the balance between the loss terms. Therefore, we use $\lambda^{\mathcal{L}}_{\mathrm{ADD}\text{-}\mathrm{S}}=30.0$ as it provides the best trade-off across the evaluated metrics.\\
\noindent\textbf{Pose loss analysis.}
Tab.~\ref{tab:ablation}(c) evaluates the contribution of individual pose-loss
terms by setting one term to zero while retaining all remaining weights.
Removing the ADD-S loss causes the largest drop, from 64.6 to 33.1 ADD(-S) AUC and from 85.9 to 74.9 ADD-S AUC, confirming that it is essential for
accurate pose estimation. Removing the depth translation loss or center loss also degrades performance, reducing ADD(-S) AUC to 59.9 and 62.2,
respectively. Overall, all pose-loss components contribute to performance, with the ADD-S term having the strongest impact.\\

\begin{table}[htb]
\centering
\small
\setlength{\tabcolsep}{5pt}
\begin{tabular}{llcc}
\toprule
& \textbf{Setting} & \makecell{\textbf{AUC of}\textbf{ADD(-S)} $\uparrow$}
                   & \makecell{\textbf{AUC of}\textbf{ADD-S} $\uparrow$} \\
\midrule
\multirow{4}{*}{\makecell[l]{(a)   $\lambda^{\mathcal{C}}_{\mathrm{cntr}}$}}
 & $0.0$ & 63.1 & 84.5 \\
 & $1.0$ & 62.9 & 84.9 \\
 & $2.5$ & 63.3 & 85.3 \\
 & \cellcolor{yellow!50}$\mathbf{5.0}$
        & \cellcolor{yellow!50}\textbf{64.6} & \cellcolor{yellow!50}\textbf{85.9} \\
\midrule
\multirow{5}{*}{\makecell[l]{(b)  $\lambda^{\mathcal{L}}_{\mathrm{ADD}\text{-}\mathrm{S}}$}}
 & $10.0$ & 57.1 & 79.5 \\
 & $20.0$ & 58.5 & 81.0 \\
 & \cellcolor{yellow!50}$\mathbf{30.0}$ 
        & \cellcolor{yellow!50}\textbf{64.6} & \cellcolor{yellow!50}\textbf{85.9} \\
 & $50.0$ & 59.4 & 83.1 \\
 & $75.0$ & 57.5 & 78.6 \\
\midrule
\multirow{2}{*}{\makecell[l]{(c) Ablated\\ loss weights}}
 & $\lambda^{\mathcal{L}}_{t_z} = 0.0 $ & 59.9 & 84.4 \\
 & $\lambda^{\mathcal{L}}_{\mathrm{cntr}} = 0.0$ & 60.7 & 82.6 \\
  & $\lambda^{\mathcal{L}}_{\mathrm{ADD-S}} = 0.0$ & 33.1 & 74.9 \\
\bottomrule
\end{tabular}
\caption{Sensitivity of pose accuracy to the matching-cost (a) and loss weights (b,c) on YCB-V. One factor is varied at a time; all remaining weights are kept at the configuration described in Sec.~\ref{sec:methodology}.}\label{tab:ablation}
\end{table}
\noindent\textbf{Runtime analysis.} Tab.~\ref{tab:runtime} reports the measured inference latency of the proposed model variants on a consumer-grade NVIDIA RTX 4090 GPU. TinyDETR-Pose is designed to address the efficiency requirements of edge deployment. It achieves a per-frame latency of approximately $7\mathrm{ms}$ in a PyTorch runtime environment. To further assess the edge-deployment capability of our method, we deploy our model on an NVIDIA Jetson Nano using TensorRT-optimized weights. The optimized model attains an inference latency of approximately $4.5\mathrm{ms}$ per frame while largely preserving its pose-estimation accuracy, demonstrating the suitability of our approach for real-time edge applications.  TinyDETR-Pose achieves a comparable AUC of ADD-S while running $\sim$\,\textbf{59\%}
lower latency than any other model evaluated on that metric; when additionally deployed through the TensorRT runtime, latency drops to
4.5\,ms/frame -- a $\sim36\%$ reduction over our own PyTorch runtime, and a \textbf{60.7\%} reduction relative to the fastest prior single-stage method (RACE-6D~\cite{ha2026race6d}, 12.0\,ms), without a notable loss in performance.
% ── Runtime table ──────────────────────────────────────────────────────────────
\begin{table}[htb]
\centering
\setlength{\tabcolsep}{5pt}
\renewcommand{\arraystretch}{1.2}
\scalebox{0.9}{%
\begin{tabular}{lcccc}
\toprule
\textbf{Method} & 
\textbf{Backbone} & 
\makecell{\textbf{Num.}\\\textbf{Decoders}} &
\textbf{ms/frame} $\downarrow$ & 
\makecell{\textbf{AUC of}\\\textbf{ADD-S} $\uparrow$} \\
\midrule
CosyPose$^\dagger$~\cite{cosypose}    & EfficientNet-B3  & -- & $395$  & 89.8  \\
GDR-Net$^\ddagger$~\cite{gdrnet}      & ResNet-34        & -- & 65     & 89.1  \\

YOLO-6D-Pose (s)$^\ast$~\cite{yolox6d} & CSPDarkNet-s    & -- & 16.9   & --    \\
T6D-Direct~\cite{T6DDirect}           & ResNet-50        & 6  & 17     & 86.2  \\
YOLOPoseV2~\cite{YOLOPoseV2}          & ResNet-50        & 6  & 17     & 90.1  \\
YOLOPoseV2-A~\cite{YOLOPoseV2}        & ResNet-50        & 6  & 22     & 91.2  \\
PoET~\cite{poet}                      & Scaled-YOLOv4    & 5  & --     & 87.1  \\
RACE-6D~\cite{ha2026race6d}             & ResNet-50    & --  & 12 & --  \\
\midrule
\rowcolor{gray!12}
\textbf{TinyDETR-Pose} (ours)              & ViT-Tiny         & \textbf{3} & \textbf{~7}   & \textbf{85.9} \\
\rowcolor{gray!12}
\textbf{TinyDETR-Pose (TRT)} (ours)        & ViT-Tiny         & \textbf{3} & \textbf{~4.5} & \textbf{85.3} \\
\bottomrule
\end{tabular}%
}
\caption{Runtime comparison on YCB-V. $\dagger$: refinement-based,
$\ddagger$: direct-regression, $\ast$: NMS post-processing; all others are end-to-end. If reported for DETR-based methods, decoder layers are reported. Latency is given in ms/frame. TinyDETR-Pose maintains comparable AUC of ADD-S  with
substantially lower latency, further improved by TensorRT deployment on the NVIDIA Jetson Nano.}
\label{tab:runtime}
\end{table}

\begin{table}[ht]
\centering
\setlength{\tabcolsep}{4pt}
\renewcommand{\arraystretch}{1.2}
\resizebox{\columnwidth}{!}{%
\begin{tabular}{l ccccc | ccccc}
\toprule
\textbf{Metric} & \multicolumn{5}{c}{\textbf{AUC of ADD-S}} & \multicolumn{5}{c}{\textbf{AUC of ADD(-S)}} \\
\cmidrule(lr){2-6} \cmidrule(lr){7-11}
\textbf{Object} 
& \makecell{\textbf{TinyDETR-}\\\textbf{Pose} (ours)} & \textbf{PoET}~\cite{poet} & \makecell{\textbf{YOLO}\\\textbf{Pose}}~\cite{YOLOPoseV2} & \makecell{\textbf{YOLO}\\\textbf{Pose-A}~\cite{YOLOPoseV2}} & \textbf{T6D}~\cite{T6DDirect}
& \makecell{\textbf{TinyDETR-}\\\textbf{Pose} (ours)} & \textbf{PoET}~\cite{poet} & \makecell{\textbf{YOLO}\\\textbf{PoseV2}~\cite{YOLOPoseV2}} & \makecell{\textbf{YOLO}\\\textbf{PoseV2-A}~\cite{YOLOPoseV2}} & \textbf{T6D}~\cite{T6DDirect} \\
\midrule
master chef can         & \cellcolor{blue!15}90.3 & 88.4 & 91.3 & \cellcolor{yellow!50}\textbf{91.7} & 91.1 & 64.6 & -  & 64.0 & \cellcolor{yellow!50}\textbf{71.3} & 61.5 \\
cracker box             & 78.5 & 80.5 & 86.8 & \cellcolor{yellow!50}\textbf{92.0} & 86.6 & 6.5  & -  & 77.9 & \cellcolor{yellow!50}\textbf{83.3} & 76.3 \\
sugar box               & 83.6 & 92.4 & \cellcolor{yellow!50}\textbf{92.6} & 91.5 & 90.3 & 67.1 & -  & \cellcolor{yellow!50}\textbf{87.3} & 83.6 & 81.8 \\
tomato soup can         & \cellcolor{blue!15}89.7 & \cellcolor{yellow!50}\textbf{91.4} & 90.5 & 87.8 & 88.9 & 64.7 & -  & \cellcolor{yellow!50}\textbf{77.8} & 72.9 & 72.0 \\
mustard bottle          & 88.8 & 91.7 & 93.6 & \cellcolor{yellow!50}\textbf{96.6} & 94.7 & 62.8 & -  & 87.9 & \cellcolor{yellow!50}\textbf{93.4} & 85.7 \\
tuna fish can           & \cellcolor{blue!15}93.3 & 90.4 & 94.3 & \cellcolor{yellow!50}\textbf{94.9} & 92.2 & 50.3 & -  & \cellcolor{yellow!50}\textbf{74.4} & 70.5 & 59.0 \\
pudding box             & 85.0 & 89.0 & 92.3 & \cellcolor{yellow!50}\textbf{92.6} & 85.1 & 61.4 & -  & \cellcolor{yellow!50}\textbf{87.9} & 87.0 & 72.7 \\
gelatin box             & 88.3 & 91.7 & 90.1 & \cellcolor{yellow!50}\textbf{92.2} & 86.9 & 73.1 & -  & 83.4 & \cellcolor{yellow!50}\textbf{85.7} & 74.4 \\
potted meat can         & \cellcolor{blue!15}90.1 &\cellcolor{yellow!50} \textbf{91.2} & 85.8 & 85.0 & 83.5 & 62.7 & -  &\cellcolor{yellow!50}\textbf{ 76.7} & 71.4 & 67.8 \\
banana                  & 87.7 & 89.5 & 95.0 & \cellcolor{yellow!50}\textbf{95.8} & 93.8 & 46.1 & -  & 88.2 & \cellcolor{yellow!50}\textbf{90.0} & 87.4 \\
pitcher base            & 90.4 & 91.7 & 93.6 & \cellcolor{yellow!50}\textbf{95.2} & 92.3 & 68.2 & -  & 88.5 & \cellcolor{yellow!50}\textbf{90.8} & 84.5 \\
bleach cleanser         & 79.5 &\cellcolor{yellow!50}\textbf{85.4} & 85.3 & 83.1 & 83.0 & 53.2 & -  & \cellcolor{yellow!50}\textbf{73.0} & 70.8 & 65.0 \\
bowl$^{*}$              & 83.8 & 90.5 & 92.3 & \cellcolor{yellow!50}\textbf{93.4} & 91.6 & 83.8 & -  & 92.3 & \cellcolor{yellow!50}\textbf{93.4} & 91.6 \\
mug                     & 91.0 & 91.4 & 84.9 & \cellcolor{yellow!50}\textbf{95.5} & 89.8 & 70.0 & -  & 69.6 & \cellcolor{yellow!50}\textbf{90.0} & 72.1 \\
power drill             & 86.3 & 88.8 & \cellcolor{yellow!50}\textbf{92.6} & 92.5 & 88.8 & 66.1 & -  & \cellcolor{yellow!50}\textbf{ 86.1} & 85.2 & 77.7 \\
wood block$^{*}$        & 83.9 & 75.7 & 84.3 & \cellcolor{yellow!50}\textbf{93.0} & 90.7 & 83.9 & -  & 84.3 & \cellcolor{yellow!50}\textbf{93.0} & 90.7 \\
scissors                & 80.8 & 75.2 & \cellcolor{yellow!50}\textbf{93.3} & 80.9 & 83.0 & 51.0 & -  & \cellcolor{yellow!50}\textbf{87.0} & 71.2 & 59.7 \\
large marker            & \cellcolor{yellow!50}\textbf{86.5} & 81.2 & 84.9 & 85.2 & 74.9 & \cellcolor{blue!15}74.6 & -  &76.6 & \cellcolor{yellow!50}\textbf{77.0} & 63.9 \\
large clamp$^{*}$       & 80.3 & 88.6 & 92.0 & \cellcolor{yellow!50}\textbf{94.7} & 78.3 & 80.3 & -  & 92.0 & \cellcolor{yellow!50}\textbf{94.7} & 78.3 \\
extra large clamp$^{*}$ & 74.1 & 83.5 & \cellcolor{yellow!50}\textbf{88.9} & 80.7 & 54.7 & 74.1 & -  & \cellcolor{yellow!50}\textbf{88.9} & 80.7 & 54.7 \\
foam brick$^{*}$        & \cellcolor{blue!15}92.1 & 81.3 & 90.7 & \cellcolor{yellow!50}\textbf{93.8} & 89.9 & \cellcolor{blue!15}92.1 & -  & 90.7 & \cellcolor{yellow!50}\textbf{93.8} & 89.9 \\
\midrule
\rowcolor{gray!12}
\textbf{MEAN}           & 85.9 & 87.1 & 90.1 & \textbf{91.2} & 86.2 & 64.6 & -- & 82.6 & \textbf{83.3} & 74.6 \\
\bottomrule
\end{tabular}%
}
\caption{Per-object results on YCB-V. $^{*}$ marks symmetric objects. We compare single-stage end-to-end DETR-based methods using AUC of ADD-S and ADD(-S), where available. Results within $3.0$ AUC points of the state-of-the-art are highlighted in \colorbox{blue!15}{blue}.}
\label{tab:ycbv_per_object}
\end{table}
\noindent\textbf{Per object analysis.}
Tab.~\ref{tab:ycbv_per_object} provides a detailed per-object comparison on YCB-V benchmark dataset~\cite{posecnn}. Overall, TinyDETR-Pose achieves competitive ADD-S performance on several objects despite its lightweight design, and obtains the best result for the \emph{large marker}. In addition, our model performs on par with the strongest competing methods for objects such as \emph{master chef can}, \emph{tomato soup can}, \emph{tuna fish can}, \emph{potted meat can}, \emph{foam brick} and \emph{large marker}. However, the ADD(-S) results reveal larger performance gaps for some asymmetric objects, most notably the \emph{cracker box}, \emph{banana} and \emph{scissors}, indicating difficulty with fine-grained rotational alignment for several asymmetric objects. We attribute this primarily to heavy occlusions in these instances, compounded by the limited refinement capacity of our shallow three-layer decoder. Increasing decoder depth or adding occlusion-aware reasoning could address this gap.

\section{Conclusion}\label{sec:conclusion}
We presented \mbox{\textbf{TinyDETR-Pose}}, a lightweight end-to-end framework
for single-stage 6D object pose estimation. By building on LW-DETR and
formulating joint object detection and pose estimation as a set-prediction
problem, TinyDETR-Pose directly regresses rotation, projected object center, and
metric depth from decoder queries, without requiring PnP, NMS, or iterative
pose refinement. The proposed symmetry-agnostic ADD-S supervision further
allows a unified training objective for both symmetric and asymmetric objects,
removing the need for explicit symmetry labels or object-specific pose losses.
Experiments on YCB-V show that TinyDETR-Pose achieves competitive pose
accuracy while being substantially more efficient than existing DETR-based
single-stage pose-estimation approaches. In particular, our model obtains an
AUC of ADD-S of \textbf{85.9} while requiring up to
$72.7 \%$ fewer parameters than other DETR-based methods. Moreover,
TinyDETR-Pose runs in real time and reaches an inference latency of only
${\sim}4.5$\,ms per frame on an NVIDIA Jetson Nano using TensorRT. These results
demonstrate that accurate transformer-based 6D pose estimation can be made
practical for resource-constrained edge deployment.\\ 
\noindent\textbf{Future Work \& Limitations.}
Although our ADD-S AUC is within 4.2--5.3 points of the strongest single-stage
transformer methods, our ADD(-S) AUC is 19.8 points lower than YOLO-6D-Pose (s)
and 18.7 points lower than YOLOPoseV2-A. We attribute this gap to two main
factors. First, a loss--metric mismatch: our model is trained with the
symmetry-agnostic ADD-S objective, which computes the nearest-neighbor
distance between transformed model points and may not sufficiently penalize
rotational ambiguities of asymmetric objects, such as flips of box-like
shapes, whereas ADD(-S) explicitly penalizes these errors. The per-object
results in Tab.~\ref{tab:ycbv_per_object} support this observation, as the largest gaps occur for
asymmetric, box- or plane-dominated objects such as the cracker box and
scissors, while symmetric objects and cans are largely unaffected. Second,
limited model capacity and the lack of a refinement stage: TinyDETR-Pose uses
a three-layer decoder and a ViT-Tiny backbone ($70.2-72.7\%$ fewer
parameters than larger alternatives\cite{YOLOPoseV1,YOLOPoseV2}) and does not include iterative or
keypoint-conditioned pose refinement~\cite{ha2026race6d}. We therefore position TinyDETR-Pose as an
efficiency-oriented approach rather than one targeting state-of-the-art pose
accuracy.
Despite its efficiency and strong real-time performance, several directions
remain open. The evaluation should be extended beyond YCB-V to benchmarks such
as LM-O~\cite{linemod} and industrial datasets with stronger occlusions,
textureless objects, and domain-specific geometries. Pose-specific
augmentations, as used in YOLO-6D-Pose~\cite{yolox6d}, could improve
robustness to occlusion, truncation, and illumination changes. The Hungarian
matcher, currently symmetry-safe via class and 2D spatial cues, could be
extended with occlusion- or pose-aware costs without destabilizing symmetric objects. Because some implementations use fewer queries~\cite{YOLOPoseV1,T6DDirect}, $N$ could be reduced to lower computational cost. Recently presented rotation representations such as SARR~\cite{SARR} may improve accuracy on ambiguous or partially symmetric objects, and edge-oriented optimizations (quantization, pruning, distillation, improved TensorRT deployment) could further increase runtime efficiency. Finally, systematic analysis of pretrained weights and scaling behavior—following the strong pretraining benefits seen in DETR-based 2D detectors—could improve representation quality and close the remaining accuracy gap while preserving TinyDETR-Pose's efficiency advantages.

\section*{Acknowledgements}
This project was funded in part by the European Union’s Horizon Europe research and innovation programme under grant agreement No.\ 101120726

\bibliographystyle{unsrt}
\bibliography{main}

\end{document}